\documentclass[conference,compsoc]{IEEEtran}
\ifCLASSOPTIONcompsoc
  \usepackage[nocompress]{cite}
\else
  \usepackage{cite}
\fi
\ifCLASSINFOpdf
\else
\fi
\usepackage{amsmath}
\usepackage{amssymb}
\usepackage{amsthm}
\usepackage{mathtools}
\usepackage{bm}
\usepackage{mathrsfs}
\usepackage{cases}
\usepackage{esvect}
\usepackage{siunitx}
\usepackage{algorithm}
\usepackage{booktabs} 
\usepackage{tabularx} 
\usepackage{graphicx}
\pdfoutput=1
\begin{document}
\pagestyle{plain}
\title{Utilizing Machine Learning and 3D Neuroimaging to Predict Hearing Loss: A Comparative Analysis of Dimensionality Reduction and Regression Techniques}
\author{\IEEEauthorblockN{Trinath Sai Subhash Reddy Pittala}
	\IEEEauthorblockA{School of Computing\\
		Clemson University\\
	tpittal@g.clemson.edu}
	\and
	\IEEEauthorblockN{Uma Maheswara R Meleti}
	\IEEEauthorblockA{School of Computing\\
		Clemson University\\
	umeleti@g.clemson.edu}
	\and
	\IEEEauthorblockN{Manasa Thatipamula}
	\IEEEauthorblockA{School of Computing\\
		Clemson University\\
        mthatip@g.clemson.edu}}

\maketitle
\begin{abstract}
	
	In this project, we have explored machine learning approaches for predicting hearing loss thresholds on the brain's gray matter 3D images. We have solved the problem statement in two phases. In the first phase, we used a 3D CNN model to reduce high-dimensional input into latent space and decode it into an original image to represent the input in rich feature space. In the second phase, we utilized this model to reduce input into rich features and used these features to train standard machine learning models for predicting hearing thresholds. We have experimented with autoencoders and variational autoencoders in the first phase for dimensionality reduction and explored random forest, XGBoost and multi-layer perceptron for regressing the thresholds. We split the given data set into training and testing sets and achieved an 8.80 range and 22.57 range for PT500 and PT4000 on the test set, respectively. We got the lowest RMSE using multi-layer perceptron among the other models.
	
	Our approach leverages the unique capabilities of VAEs to capture complex, non-linear relationships within high-dimensional neuroimaging data. We rigorously evaluated the models using various metrics, focusing on the root mean squared error (RMSE). The results highlight the efficacy of the multi-layer neural network model, which outperformed other techniques in terms of accuracy. This project advances the application of data mining in medical diagnostics and enhances our understanding of age-related hearing loss through innovative machine-learning frameworks.
\end{abstract}

\IEEEpeerreviewmaketitle

\section{Introduction}

Age-related hearing loss (ARHL), also known as presbycusis, is an increasingly prevalent condition that affects a significant portion of the aging population. Characterized by the gradual loss of hearing capabilities, ARHL poses substantial challenges to interpersonal communication and overall quality of life. Traditional methods of diagnosing and evaluating ARHL rely on auditory tests and essential medical imaging; however, these methods often fall short of identifying the underlying neuroanatomical changes associated with the condition.

Recent medical imaging and machine learning advancements present new opportunities to enhance our understanding of ARHL. Specifically, brain magnetic resonance imaging (MRI) offers detailed insights into the brain's structural integrity. At the same time, sophisticated data mining techniques enable the analysis of complex datasets to reveal patterns not immediately apparent to human observers. This project leverages these technologies to bridge the gap between neuroanatomical changes and auditory function, employing a dataset that combines gray matter images with auditory threshold measurements.

Our research employs advanced machine learning techniques, including variational autoencoders (VAEs) and multi-layer neural networks, to analyze the relationship between brain structure and hearing loss. These models are particularly suited for this task because they can handle high-dimensional data and learn deep representations that capture the subtle nuances of brain morphology related to ARHL. By training these models on a robust dataset from clinical settings, this project aims to predict hearing thresholds more accurately and identify specific brain regions that correlate with hearing decline.

Through this approach, the project seeks to contribute to the broader field of medical diagnostics by providing a more nuanced understanding of ARHL, paving the way for improved diagnostic accuracy and potentially informing targeted intervention strategies that could mitigate hearing loss impacts in older people.

\section{Literature Review}

Research into the correlation between brain morphology and sensory loss, particularly hearing, has been extensively documented. Studies leveraging MRI technology have identified significant brain structure changes in individuals with hearing impairment compared to those without. Deep learning has emerged as a powerful tool for deciphering these complex patterns, providing predictive capabilities that surpass traditional statistical methods. This project builds on these foundations by applying advanced machine learning models to predict auditory thresholds from MRI-derived brain images.

\begin{figure*}[htbp]
    \centering
    \includegraphics[width=\textwidth]{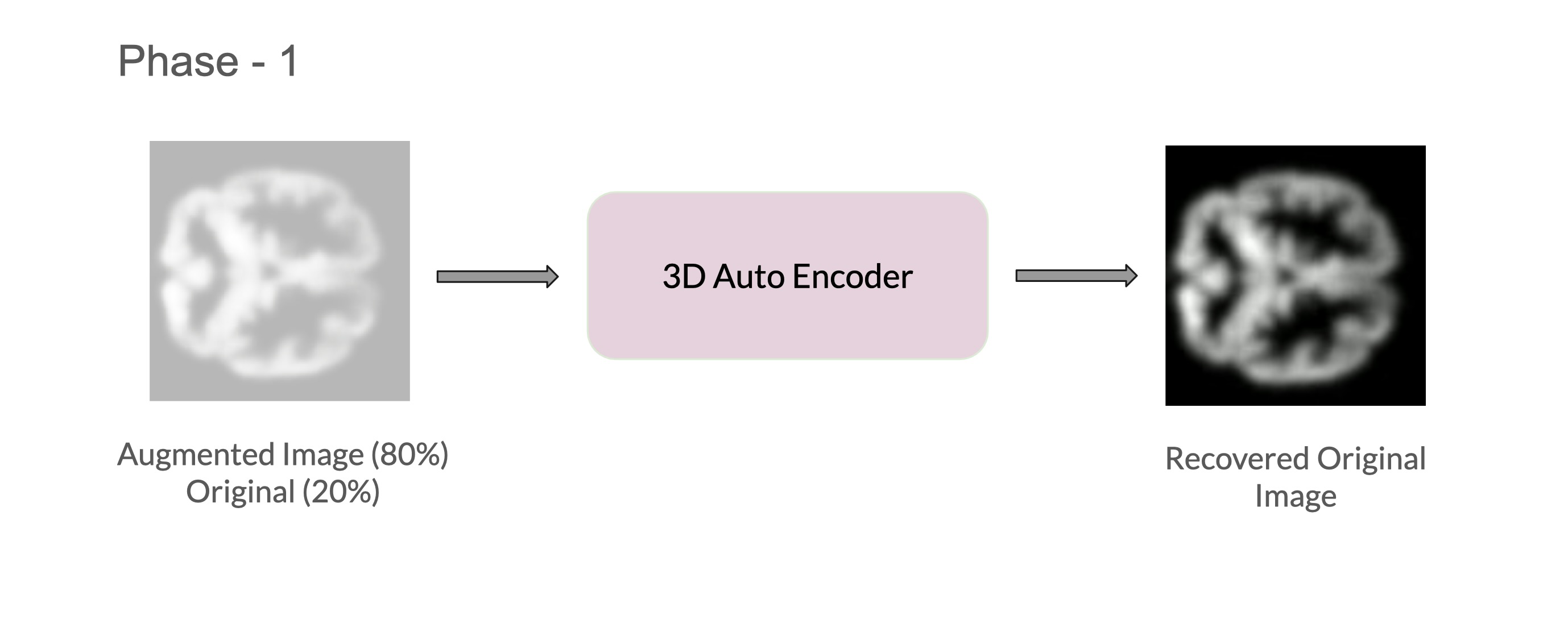}  % Adjust this value to change the width
    \caption{}
    \label{phase1}
\end{figure*}

\section{Methodology}

\subsection{Data Acquisition:} The primary dataset for this project consisted of 171 gray matter images and 137 slices of human brain scans obtained from MRI scans. Each image was meticulously labeled with corresponding hearing thresholds for 500 Hz and 4000 Hz frequencies. This dual-frequency labeling allowed for a nuanced analysis of hearing capabilities across different acoustic spectrums, facilitating a more comprehensive training and validation process for our predictive models. By leveraging such a specifically curated dataset, we ensured that our models could be trained with high fidelity to actual clinical conditions, improving the relevance and applicability of our findings.

\subsection{Data Preparation} A series of sophisticated data augmentation techniques were applied to the original MRI scans in preparation for the training process. This was done to simulate a variety of real-world variations that could affect the image quality and integrity, thus ensuring our model's robustness and generalizability across different medical imaging environments.

\subsubsection{Random Noise Injection:} This technique involved the addition of synthetic noise to the images, mimicking the effect of various interferences found in typical clinical MRI scans. By training our models to recognize and adjust for these perturbations, we aimed to enhance their performance in less-than-ideal imaging conditions.

\subsubsection{Gamma Adjustments:} By altering the gamma values of the images, we modified the luminance, helping our models to remain effective under different lighting and exposure conditions. This is crucial for ensuring the models' performance in diverse clinical settings and with varying MRI machine calibrations.

\subsubsection{Blurring:} Applying a Gaussian blur mimicked the effect of slight focus variations that often occur in real-world scans. This method trained the models to extract relevant features from images that are not perfectly sharp, a common issue in medical imaging.

\subsubsection{Random Bias Field} MRI scans can often exhibit intensity inhomogeneities known as bias fields, which can affect the uniformity of the image. This augmentation introduces synthetic bias fields into the images, ensuring the model can handle such inconsistencies without compromising the diagnostic accuracy.

\subsubsection{Random Spike} Spike artifacts in MRI can occur due to hardware malfunctions or power fluctuations, resulting in sharp spikes in the signal intensity. Including this noise in the training, data helps prepare the model to process and analyze images even with potential spike artifacts.

\subsubsection{Random Ghosting} Ghosting artifacts are repetitive motion artifacts that appear as additional images or "ghosts" superimposed on the original image. This can happen due to patient movement or issues with the MRI scanner's phase encoding. Training with ghosted images ensures the model can differentiate between true anatomical structures and artifacts.

\subsubsection{Random Motion} To simulate the effects of patient movement during the scan, random motion artifacts are introduced. These can vary from slight to severe, mimicking the blurring and distortion such movements can cause. Motion artifacts are crucial for training robust models that operate effectively in less-than-ideal scanning conditions.

\begin{figure*}[htbp]
	\centering
	\includegraphics[width=\textwidth]{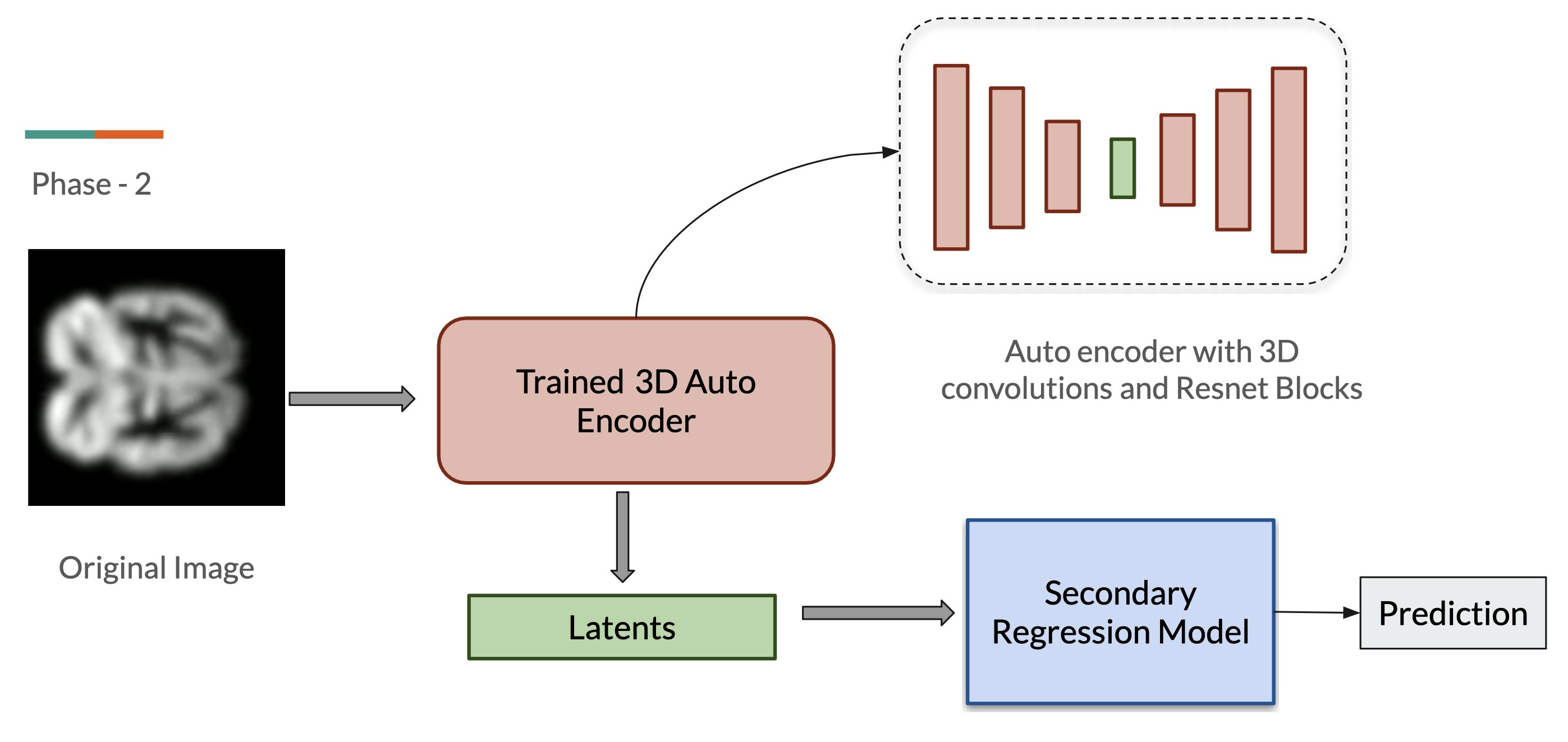}
	\caption{}
	\label{phase2}
\end{figure*}

\subsection{Modeling Approach} The modeling approach was structured in two distinct phases.

\subsubsection{Dimensionality Reduction}

\begin{itemize}
	\item {\textbf{Autoencoders}}: The first phase of our methodology involved the use of 3D convolutional autoencoders Figure~\ref{phase1}. The main purpose of this step was to compress the vast data in the MRI scans into a more manageable form without significant loss of critical information. This reduction was crucial for simplifying the computational requirements and distilling the essential features most indicative of hearing thresholds. By training the autoencoders to reconstruct the original input from the compressed encoded version, we effectively taught the models to capture and prioritize the most relevant features in the data.
\end{itemize}

\subsubsection{Regression Modeling}

\begin{itemize}
	\item {\textbf{Secondary Models}}: The latent representations generated by the autoencoders were then utilized as inputs for the project's second phase, where more focused predictive modeling took place Figure~\ref{phase2}. We used several advanced machine learning algorithms in this phase, including Random Forest, XGBoost, and Multi Neural Networks (MNN). Each model was chosen for its unique strengths in handling complex, non-linear data relationships, typical in medical imaging data. The Random Forest model provided a robust and reliable baseline due to its ensemble approach. XGBoost offered highly efficient and scalable processing, making it ideal for dealing with the reduced yet still complex data from the autoencoders. Lastly, the MNN was used for its ability to capture patterns through deep learning, which is essential for accurately predicting medical outcomes from imaging data.
\end{itemize}

\section{Models}
\subsection{Autoencoder} 
\begin{figure*}[htbp]
	\centering
	\includegraphics[width=\textwidth]{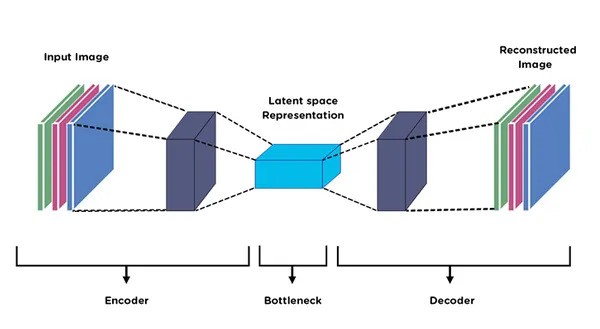}
	\caption{}
	\label{ae}
\end{figure*}

Autoencoders are neural networks designed for unsupervised learning of efficient codings, primarily used for dimensionality reduction Figure~\ref{ae}. The architecture consists of an encoder that compresses the data, a bottleneck code layer, and a decoder that reconstructs the original input. Our project used a 3D convolutional autoencoder to manage the spatial hierarchy inherent in MRI images. The encoder compressed each high-dimensional brain MRI image into a lower-dimensional latent space, capturing essential features and minimizing information loss. This was crucial for efficiently handling the vast data typically involved in medical imaging and enhanced the computational efficiency of our predictive modeling.

\subsection{Variational Autoencoder (VAE)} 
\begin{figure}[htbp]
	\centering
	\includegraphics[width=0.5\textwidth]{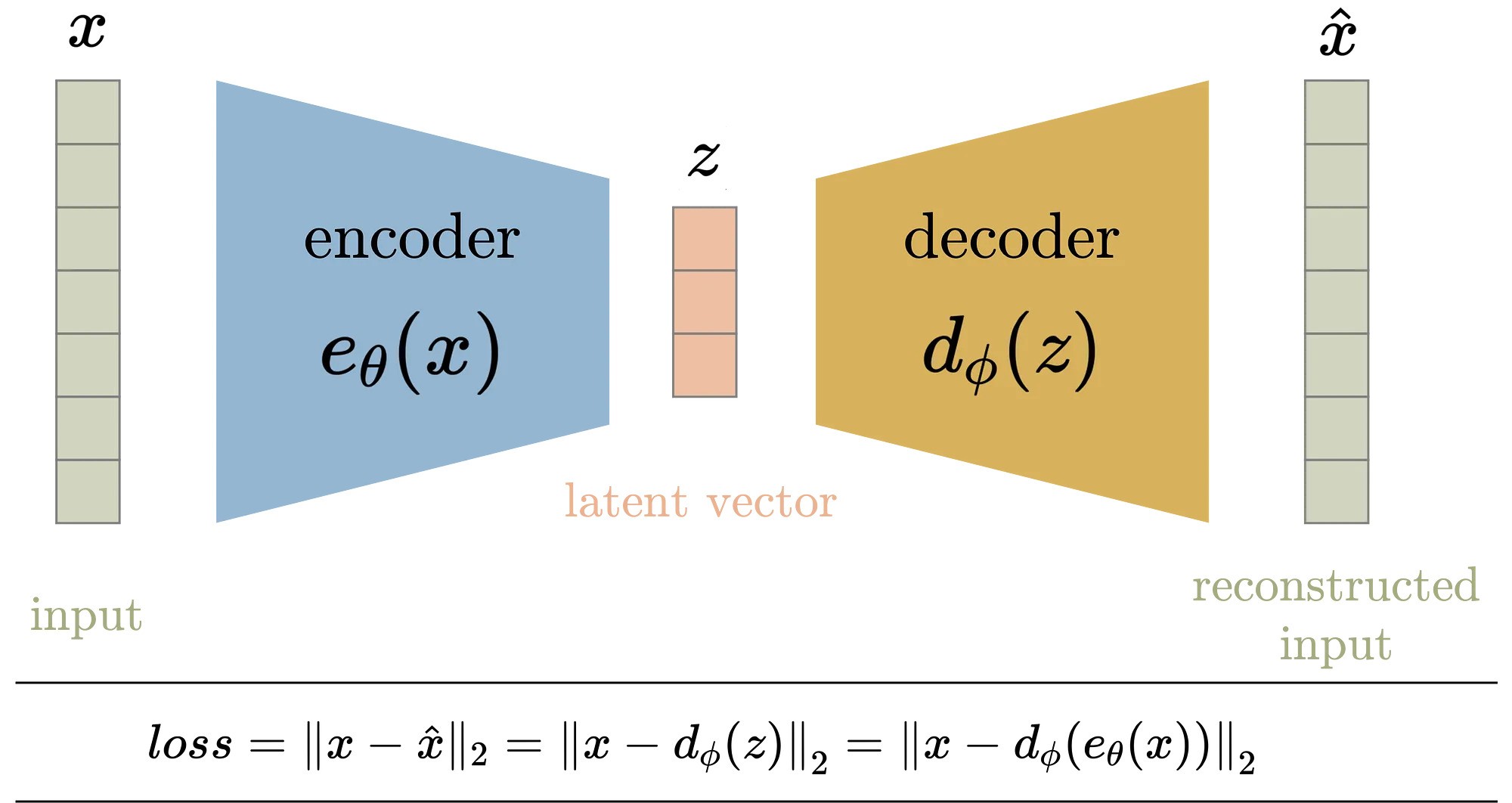}
	\caption{}
	\label{vae}
\end{figure}

Variational Autoencoders (VAEs) extend the capabilities of standard autoencoders by introducing a probabilistic approach to encoding inputs Figure~\ref{vae}. Instead of encoding an input as a single point, VAEs encode inputs as distributions—characterized by mean and variance—over latent space. This enables the generation of new data points and helps understand the underlying distribution of data features. In our study, the VAE was instrumental in reducing data dimensionality and modeling the distribution in the latent space, which provided valuable insights into how different features of brain anatomy are associated with hearing thresholds, thereby exploring the biological aspects of age-related hearing loss.

\subsection{XGBoost} XGBoost (Extreme Gradient Boosting) is an advanced implementation of gradient boosting algorithms known for their efficiency and effectiveness in handling large-scale and sparse data. It operates under the Gradient Boosting framework to provide a robust solution to various data science challenges. Our project utilized XGBoost to process latent features extracted from the autoencoder to predict hearing thresholds. Its advanced regularized boosting techniques were crucial in preventing overfitting, making it exceptionally suitable for our regression tasks and enhancing the predictive accuracy of the hearing thresholds.

\begin{table*}[ht]
	\centering
	\caption{Performance of Different Models on Prediction Task}
	\label{tab:model_performance}
	\begin{tabular}{|l|c|c|}
		\hline
		\textbf{Model}              & \textbf{RMSE of PT500 (dB)} & \textbf{RMSE of PT4000 (dB)} \\ \hline
		Random Forest               & 9.2287                      & 22.8441                      \\ \hline
		XGBoost                     & 9.5331                      & 24.3296                      \\ \hline
		MNN                         & 8.8057                      & 22.5222                      \\ \hline
		Ensemble of MNN and XGBoost & 9.4629                      & 25.9908                      \\ \hline
	\end{tabular}
        \label{table}
\end{table*}

\subsection{Random Forest} Random Forest is an ensemble learning method that constructs multiple decision trees during training and outputs the mode or mean prediction of these trees for classification and regression tasks, respectively. It is renowned for its robustness against overfitting, a critical advantage in medical applications where prediction accuracy can significantly impact outcomes. In our project, Random Forest was used to average predictions across many trees, each trained on different data subsets, thus ensuring more stable and reliable predictions for hearing thresholds from MRI data.

\subsection{Multi Neural Network (MNN)} Multi Neural Networks (MNN) involve multiple layers of neurons, each fully connected to the next, with non-linear activation functions used at each layer except the input nodes. This structure allows MNNs to learn complex patterns through the network's depth and breadth. In our application, the MNN, configured with two hidden layers, was adept at processing the intricate non-linear interactions typical of medical image data. This allowed it to excel in predicting hearing thresholds, outperforming simpler models.

\subsection{Ensemble Model with MNN and XGBoost} Ensemble methods combine multiple learning algorithms to achieve better predictive performance than any individual model. By leveraging the strengths of each model, ensemble methods aim to enhance prediction accuracy and robustness. In our project, the ensemble model combining MNN and XGBoost aimed to merge the detailed feature recognition ability of neural networks with the efficient structure of gradient boosting. However, this approach required careful tuning as the ensemble did not perform as expected, indicating that integrating outputs from different model types can be challenging and requires precise alignment of decision boundaries.

\section{Results}
The performance of each predictive model was assessed using the Root Mean Square Error (RMSE) metric for both PT500 and PT4000 hearing thresholds, revealing distinct outcomes for each model as shown in Table~\ref{table}. The Multi Neural Network (MNN) showcased outstanding accuracy, achieving the lowest RMSE values of 8.8057 dB for PT500 and 22.5222 dB for PT4000. This performance highlights the MNN's superior ability to effectively model the complex relationships present in MRI data.

\vspace{\baselineskip} % Adds extra line space

In comparison, the Random Forest and XGBoost models also showed commendable performance but did not match the precision of the MNN. The RMSE for Random Forest was 9.2287 dB for PT500 and 22.8441 dB for PT4000, while XGBoost recorded slightly higher errors of 9.5331 dB for PT500 and 24.3296 dB for PT4000. These results demonstrate the robustness of these models, though they were slightly less effective than the MNN in this specific application.

\vspace{\baselineskip} % Adds extra line space

However, the ensemble model that combined the strengths of MNN and XGBoost did not perform as anticipated. It yielded higher RMSEs of 9.4629 dB for PT500 and 25.9908 dB for PT4000, suggesting possible overfitting or challenges in model integration. This finding underscores the complexities of blending different advanced modeling approaches and highlights the need for careful tuning and integration strategies to harness their combined potential effectively.

\section{Conclusion}

This study highlighted the potential of various machine learning models to predict age-related hearing loss from MRI-derived gray matter images. The MNN model proved the most effective, underscoring the value of neural networks in medical image analysis. Future work could explore larger datasets and refine model architectures to enhance predictive accuracy further.\\

\end{document}